\documentclass[11pt,a4paper]{article}
\usepackage[hyperref]{emnlp2020}
\usepackage{times}
\usepackage{latexsym}
\usepackage{graphicx}
\usepackage{caption}
\usepackage{subcaption}
\usepackage{booktabs}
\usepackage{multirow}
\usepackage{multicol}
\usepackage{amsmath}
\usepackage{courier}
\usepackage{url}
\usepackage{CJKutf8}

\usepackage{microtype}

\usepackage{enumitem}
\setlist{nosep} 

\usepackage{tikz}
\usepackage{pgfplots}
\pgfplotsset{width=7.5cm,compat=1.12}
\usepgfplotslibrary{fillbetween}

\usepackage{makecell}

\usepackage{xcolor}

\aclfinalcopy 

\title{MEEP: An Open-Source Platform for Human-Human Dialog Collection and End-to-End Agent Training}

\author{Arkady Arkhangorodsky, Amittai Axelrod, Christopher Chu, Scot Fang, \\ \bf Yiqi Huang, Ajay Nagesh, Xing Shi, Boliang Zhang, Kevin Knight  \\
DiDi Labs\\
4640 Admiralty Way \\
Marina del Rey, CA 90292}

\date{}

\begin{document}

\maketitle

\begin{abstract}
We create a new task-oriented dialog platform (MEEP) where agents are given considerable freedom in terms of utterances and API calls, but are constrained to work within a push-button environment. We include facilities for collecting human-human dialog corpora, and for training automatic agents in an end-to-end fashion.  We demonstrate MEEP with a dialog assistant that lets users specify trip destinations.
\end{abstract}

\section{Introduction}

We create a new open-source platform for task-oriented dialog, called MEEP\footnote{\tt \small https://github.com/didi/MEEP} (Multi-Domain End-to-End Platform).  In MEEP, agents are given considerable freedom in terms of utterances and API calls, but are constrained to work within a push-button environment. 

MEEP includes a graphical user-interface for collecting human-human dialog corpora.  It can be adapted quickly to new task domains and task APIs.  Session logs record all user and agent choices, and are fully executable.

MEEP also includes algorithms to train dialog agents on collected corpora.  Training is end-to-end, with automatic agents deciding which button to push next, based on the dialog history and the results of previous API calls.

\section{Destination Dialog Example}

Spoken dialog provides a natural human interface to complex systems.
Popular use cases include speech-enabled cell phones, smart speakers, and automated personal assistants.
However, in spatial domains such as city navigation, the predominant user interface is graphical and touch-based.  Systems display places on maps, and users browse around those maps.  Sometimes this predominant graphical interface is not convenient, such as when a user is talking to a smart speaker.  Here, intelligent dialog must take the place of visual feedback.

We investigate a particular application:  {\em How to order a ride-hailing vehicle using a language-only dialog system.} 
A near-term use case is ordering a shared-ride car using Alexa.  A medium-term use case is stepping into an autonomous taxi and telling it where to go.  
In this application, the output of the dialog is a geographic destination, specified by latitude and longitude coordinates.
The dialog system's goal is essentially to use natural conversation to extract a latitude/longitude destination from the user. Figure~\ref{dialogs} shows sample human/human dialogs for this rich and varied domain.  

\section{Contributions}

Our contributions are:

\begin{itemize}
    \item We release an open-source dialog platform that supports human/human dialog collection and end-to-end dialog agent training.
    \item We develop a human agent interface that reduces all agent activity (API calls, utterances) to a series of clicks, yet maintains enough flexibility to satisfy users.
    \item We demonstrate the platform by creating a corpus of 490 dialogs in the domain of specifying a trip destination, and we develop and evaluate three automatic agents.
\end{itemize}

\begin{figure*}[t]

\begin{scriptsize}
    \begin{verbatim}
User:   I want to go to Starbucks on Venice Boulevard.
Agent:  There is a Starbucks in Mar Vista.
        Are you okay with that one?
User:   Is it the one across from Coffee Connection?
Agent:  Starbucks is 141.1 feet away from Coffee Connection.
        It will take us 10 minutes to get there.
        Shall we go?
User:   Great, thanks.
Agent:  Great, we are going to Starbucks.

User:   I want to go to the airport.
Agent:  Do you mean LAX?
User:   Yes.
Agent:  Which terminal? 
User:   Southwest Airlines
Agent:  That's Terminal 1.  We'll head there now.

User:   I want to go to a Mexican restaurant. It's on Lincoln Blvd.
Agent:  Is it called Clutch? 
User:   No, but it is pretty close to there.
Agent:  There is one called Casablanca which is a little further north.
User:   Oh. It's across the street from that. It might be on Rose, just off Lincoln.
Agent:  I think you mean La Cabana. Is that right? 
User:   Hey, that's right.
Agent:  Great, we'll go there now.

User:   I'd like to go to Big Bear Lake.
Agent:  Big Bear Lake is 121 miles east from our current location, and it takes around two and half hours 
        to reach there.
User:   Ok, let's go.
Agent:  Do you have a particular location in mind? 
User:   No, anywhere on the lake is fine.
Agent:  Sure, we'll go now.
    \end{verbatim}
\end{scriptsize}
\vspace*{-0.2in}
\caption{Sample human/human dialogs in which the user's goal is to specify a trip destination to an automated voice assistant or autonomous vehicle.  The agent's goals are to obtain a correct destination and to provide the user with assurance that it has gotten the destination right.  The agent's last utterance is followed by a physical action, such as dispatching or driving.}
\label{dialogs}
\end{figure*}


\section{Related Work}

In fields like machine translation, researchers dip into huge, naturally-occurring streams of input/output data that humans create for their own purposes, e.g., \cite{EsplGomis2019ParaCrawlWP}.  Researchers then build systems that seek to replicate the behavior of those humans, using simple end-to-end architectures \cite{Sutskever2014SequenceTS,Vaswani2017AttentionIA} that have superseded previous pipeline architectures \cite{Koehn2007MosesOS} and hand-built systems \cite{Hutchins1992AnIT}.

In task-oriented dialog research, this goal has not been realized. End-to-end systems are just beginning, while the ``astonishingly long-lived'' idea \cite{jurafsky18} of slots, fillers, pipelines, and hand-construction predominates.  Naturally-occurring streams of task-oriented dialogs to feed such systems do exist, e.g., in customer-service call centers.  Likewise, for the domain of specifying trip destinations, ride-sharing companies can set up human phone agents \cite{uberphone}.  However, legacy human-agent platforms often fail to capture information needed for training  dialog systems.

The Wizard-of-Oz (WOZ) framework \cite{Kelley1984AnID,fraser1991simulating,Rieser2005ACC} offers a way to control and log human agent activity.  WOZ was originally used to explore how human users might ultimately interact with an automatic dialog agent \cite{jurafsky18}, and is nowadays also used to collect training data \cite{dialog2,dialog29,dialog51}.  WOZ interfaces are less-frequently placed in the hands of real-time human agents doing real tasks, producing MT-style data suitable for end-to-end training, with no intermediate annotations.

Towards this goal, we set up a WOZ interface for human agents obtaining trip destinations from human users:

\begin{itemize}

\item For ecological validity \cite{eco}), we give minimal instructions.  The human user should tell the human agent where to go, and the agent should determine the place, confirm, and drive the user there.  This naturalistic approach contrasts with two-step paradigms, which first generate a formal dialog outline, then paraphrase that outline into English \cite{Rastogi2020TowardsSM}. 

\item Our WOZ interface collects on-line conversations between two humans, versus single-turn extensions \cite{dialog2,Eric2017KeyValueRN} or machine-partnered dialogs \cite{dialog43,dialog12,dialog72,dialog73}.

\item As we aim for end-to-end training \cite{Williams2016EndtoendLD,zhao16,Eric2017KeyValueRN,Dhingra2017EndtoEndRL,dialog8}, we do not collect intermediate annotations \cite{dialog72,dialog73,dialog26,dialog29}.  For evaluation purposes, we do collect a success/fail confirmation from the user at the end of each dialog.

\item As real users may want to go anywhere in the world, we employ a publicly-accessible backend that supports tens of millions of places, in contrast with more narrowly-bounded backend databases \cite{dialog73}.





\item In most WOZ systems, human agents are allowed to type free text, to increase expressiveness or to support studies of information presentation \cite{dialog29}. However, machine learning of human-agent language is a burden.  As in \citet{hauptmann89}, we only allow free text on the user side.  Human agents select utterances from an extensible list of templates, and they select template-filler words by clicking boxes.  Agents similarly control API calls and parmeters with box clicks.  We find that human agents can easily satisfy users' goals inside the WOZ, even though the domain is complex.

\end{itemize}

\section{Human/Human Dialog Collection}
\label{collection}


\subsection{Initial Collection}
For the trip-destination dialog task, Figure~\ref{dialogs} shows dialogs collected from two humans communicating through a chat interface.  Through this initial collection, we determine that users have two types of destinations:

\begin{itemize}
\item Precise destinations that the user can visualize in their mind.  The user might not recall the address or (frequently) the name.
\item Activities that the user wants to do (e.g., eat tacos, buy clothes, or swim).  The agent must find and suggest appropriate places.
\end{itemize}

In this initial collection, the human agent was free to use tools such as Google Maps and Yelp in a separate browser window.

We would like to perform end-to-end training on collected dialogs, in order to build an automatic agent that imitates a human agent.  However, we cannot do that with the dialogs in Figure~\ref{dialogs}, because the corpus only tracks what the agent says, not what the agent does.  In fact, the agent clicks, scrolls, and types queries into external applications such as Google Maps and Yelp.  For example, the agent might look up two places on the map and visually confirm that they are close to one another.  

\begin{figure*}[t]
\begin{center}
\includegraphics[scale=0.90]{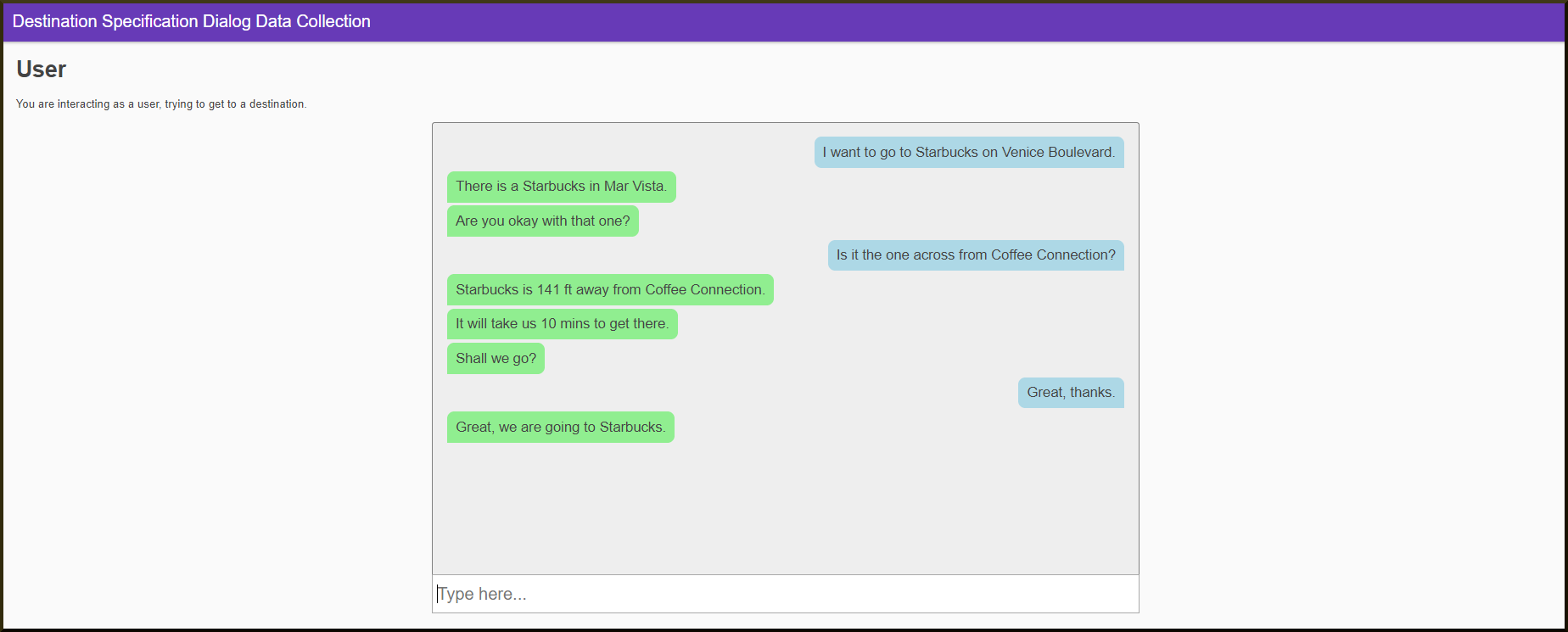} \\[0.1in]
\includegraphics[scale=0.90]{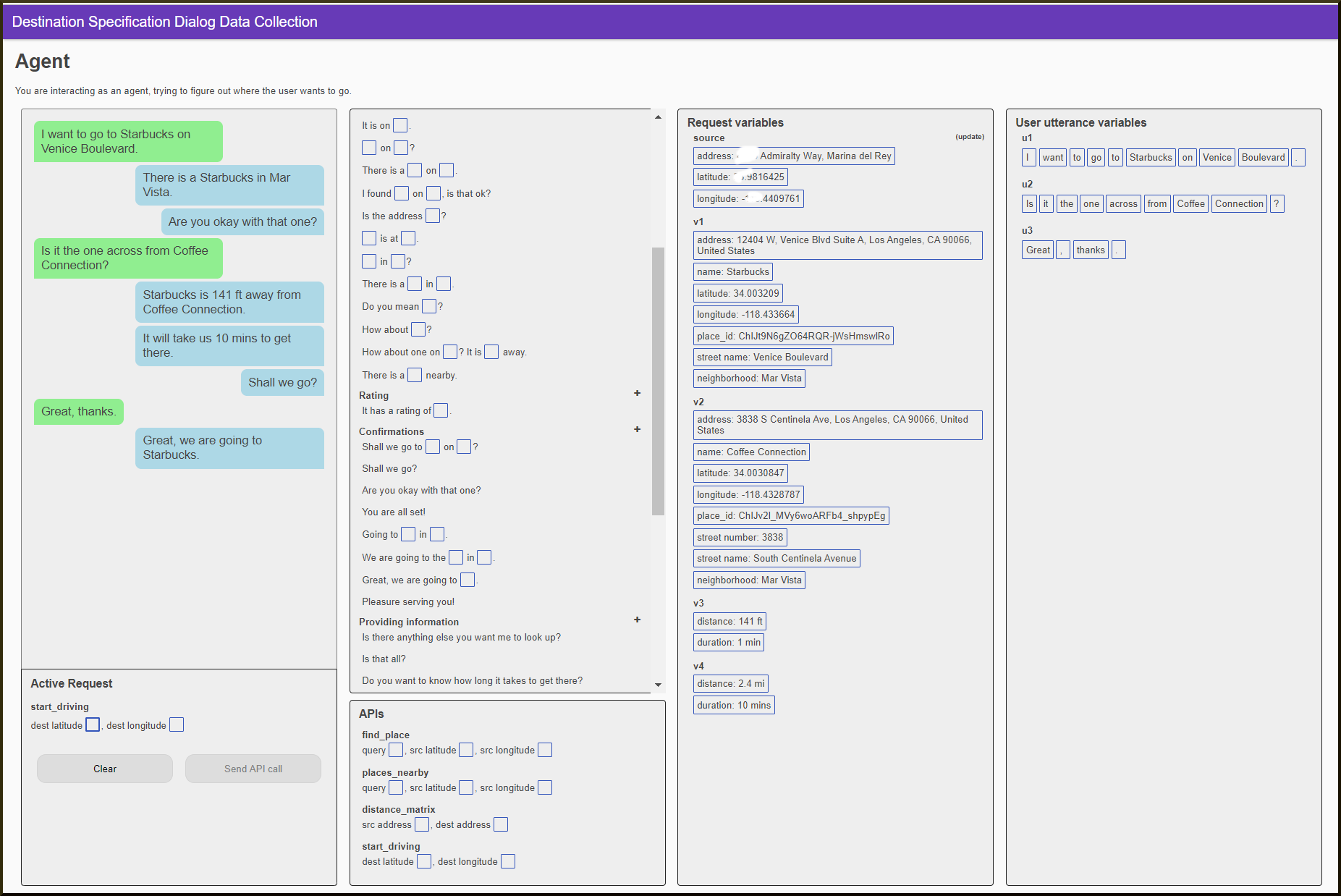}
\end{center}
\caption{Wizard-of-Oz (WOZ) interface for dialog data collection. The user sees the top interface (chat only), while the agent/operator sees the bottom interface.  The agent is only allowed to click items, such as words from user utterances, APIs, results from APIs, and templates.  That series of clicks is stored in a re-executable format.}
\label{gui}
\end{figure*}

\subsection{{\bf WOZ Interface}}  We create a WOZ human/human interface, shown in Figure~\ref{gui}.  The user side (top) allows open chat, while the agent side (bottom) constrains the agent's actions and utterances.  This platform is generic and can be adapted to other domains.

Within the agent's interface, the first panel (left) shows the chat.  The second panel shows the agent's available actions, including utterance templates (above) and API calls (below).  Results of API calls accumulate in the third panel, and tokenized user utterances are shown in the fourth panel.

Crucially, agent actions are restricted to a sequence of clicks.  For example, after the user types ``I want to go to Starbucks on Venice Boulevard'', the agent might decide to:\footnote{Note that this sequence of clicks illustrates a different agent response than shown in the figures.}

\hspace{1mm}

\begin{itemize}
    \item Click on {\tt find\_place} API from Google.  This API takes a string argument and a source lat/long.
    \begin{itemize}
    \item Click on the words ``Starbucks'', ``Venice'', and ``Boulevard'' in the fourth panel.
    \item Click on the source lat/long in the third panel.
    \item ({\em This calls the \texttt{find\_place} API in realtime and further populates the third panel with its result, collected under {\em v1}.)}
    \end{itemize}
    \item Click on the {\tt distance\_matrix} API.  This API takes two addresses, and returns the time and distance between them.
    \begin{itemize}
        \item Click on the {\em address} field of variable {\em v1} (Starbucks).
        \item Click on the {\em address} field of source.
        \item ({\em This calls the {\tt distance\_matrix} API and further populates the third panel.})
    \end{itemize}
    \item Click on the template ``{\em \{\} on \{\} is \{\} minutes away.}''  This template takes three arguments.
    \begin{itemize}
        \item Click on the {\em name} field of variable {\em v1} (``Starbucks'').
        \item Click on the {\em street name} field of variable {\em v1} (``Venice Boulevard'').
        \item Click on the {\em duration} field of variable {\em v2}
    \end{itemize}
    \item Click on the template ``{\em Shall we go?}''
\end{itemize}

\noindent
These 11 clicks result in this response, sent to the user: ``Starbucks on Venice Boulevard is 10 minutes away.  Shall we go?''

At any time, an agent can click on ``\texttt{+}'' to create a new utterance template, which is made available to all agents.  All templates shown in this paper were grown organically in this way.

{\bf Session logs.} Figure~\ref{python} shows one logging method for sessions, as fully-executable Python code.  If we run the session log file, we get a faithful replay of the session, in which the same APIs are called, and the same utterances made.


\begin{figure*}
\begin{tiny}
\begin{verbatim}
import googlemaps
from map_apis import *
gmaps = googlemaps.Client(key='xxxxxxxxxxxxxxxxxxxxxxxxxxxxxxxxxxxxxxx')

## Initial variables
source_address = 'xxx Admiralty Way, Marina del Rey'
source_latitude = 33.9816425
source_longitude = -118.4409761

# User utterance
print('User: I want to go to Starbucks on Venice Boulevard')
u1_0 = 'I'
u1_1 = 'want'
u1_2 = 'to'
u1_3 = 'go'
u1_4 = 'to'
u1_5 = 'Starbucks'
u1_6 = 'on'
u1_7 = 'Venice'
u1_8 = 'Boulevard'

# API call
api_variables = find_place(gmaps, u1_5 + " " + u1_7 + " " + u1_8, source_latitude, source_longitude)
v1_address = load_variable(api_variables, 'address')
v1_name = load_variable(api_variables, 'name')
v1_latitude = load_variable(api_variables, 'latitude')
v1_longitude = load_variable(api_variables, 'longitude')
v1_place_id = load_variable(api_variables, 'place_id')
v1_street_name = load_variable(api_variables, 'street_name')
v1_neighborhood = load_variable(api_variables, 'neighborhood')
v1_locality = load_variable(api_variables, 'locality')

# Agent utterance
print('Agent: There is a {} in {}.'.format(v1_name, v1_neighborhood))

# Agent utterance
print('Agent: Are you okay with that one?'.format())
\end{verbatim}
\end{tiny}
\caption{Stored, executable form of portion of the dialog from Figure~\ref{gui}.  Fragments of this code will replay the same actions taken by the agent.}
\label{python}
\end{figure*}

\begin{figure}
\centering
\begin{small}
\begin{tabular}{lrrr}
\toprule
\textbf{Dataset} & \textbf{Dialogs} & \textbf{Agent decisions} & \textbf{Avg.~\# of turns} \\ 
Train & 398 & 10470 & 8.4 \\
Dev & 45 & 1093 & 10.9 \\
Test & 47 & 1064 & 7.8 \\
\bottomrule
\end{tabular}
\end{small}
\caption{Statistics of collected human/human dialogs in the trip-destination domain.}
\label{data}
\end{figure}

{\bf Corpus.} Figure~\ref{data} shows that we collected 490 dialogs in the trip-destination task, including 10,000 human agent clicks.   

\section{Metrics for Automatic Agents}

Before describing how we build automatic dialog agents, we first consider how to evaluate them.

{\bf Extrinsic evaluation.}  All agents are evaluated according to successful {\em task completion} in live dialogs with graduate student users outside our lab.  After each dialog, the user is shown the target lat/long destination on a map and is asked to approve or disapprove the agent's final decision.

{\bf Intrinsic evaluation.} The intrinsic metric evaluates ``what to do/say next?'' at a finer grain.  Because trained agents are intended to mimic human agent decisions, we can test how often their individual ``click decisions'' match those of human agents.  Each train/dev/test example consists of:

\begin{itemize}
    \item The session so far.
    \item The next action to be predicted.  The action may be a API selection, a parameter selection for an API, a template selection, or a parameter selection for a template.
\end{itemize}

\noindent 
All predictions correspond to agent clicks, except for the string-query parameters to {\tt find\_place} and {\tt places\_nearby}.  For example, creating the string query ``Starbucks Venice Boulevard'' takes three clicks, but we consider it one prediction.

After replying to a user, a human agent flexibly waits for the user's next utterance.  We do not have a special button for the agent to click to wait for the user. Instead, we add a special ``click'' {\tt wait\_for\_user} between agent templates and user utterances in the collected data, so that automatic agents can learn when to wait for the user.

To better understand the agent clicks and study the system behaviors in different conditions, we categorize agent clicks into action clicks, query clicks, and parameter clicks. Action clicks are for API and template selection, such as {\tt find\_place} and {\em ``It is \{\} away.''} Query clicks are the query selection for certain API calls, for example, ``Starbucks in Venice", and parameter clicks are the parameter selections, such as latitude and longitude.


If the automatic agent does not match the human click decision, it loses an accuracy point.\footnote{String-query parameters are given partial credit via normalized character-based Levenshtein distance.}  However, it is then brought back on track, because the next example's session-so-far will end with the human decision, not the automatic agent's.  So automatic agents cannot veer off track during intrinsic evaluation, whereas in the extrinsic evaluation, automatic agents can veer very much off track.


\section{Automatic Agents}

This section describes three automatic agents: BERT-based, GPT-based, and hand-built.



\subsection{BERT-based approach}

We build a machine-learned agent based on pre-trained BERT~\cite{devlin2018bert}.\footnote{We use the HuggingFace implementation of BERT-base-cased model~\cite{wolf2019huggingfaces}: hidden size 768, intermediate size 3,072, max position embeddings 512, attention heads 12, and hidden layers 12. We consider 10 previous actions in the dialog history. We train our model on 4 Nvidia P100 GPUs. Batch size is 16 (4 per GPU). We train the model for 20 epochs and select the model checkpoint that has the highest click level accuracy on our Dev set. The training process takes about 2 hours to finish.} Action prediction predicts the API or template to invoke, given the previous user and agent context (including information retrieved from previous API calls). Query prediction locates the query for API calls, such as ``Starbucks" in the user utterance ``I want to go to Starbucks.'' Parameter prediction assigns appropriate variables retrieved by API calls to actions (APIs and templates), for example, the machine selects the API return value ``13 min" to fill in the template {\em It is \{\} away}.

We cast prediction as a sequence labeling task. We convert a partial dialog into a sequence of text (input), and we label spans of the text as action, query, or parameter (output). Specifically, we first construct a token type set that includes the APIs, templates, and two special types ``QUERY" and ``PARAMETER" which indicate query and parameter tokens. We generate the label set from this token type set using \textbf{BIO} tagging schema, where \textbf{B} means beginning of the span, \textbf{I} means inside of the span, and \textbf{O} means tokens that have no type. Then we create the \textbf{dialog context} string. We concatenate the previous user utterances, agent utterances, API calls, and variables in order, as well as adding a start of the sequence token ``\texttt{\scriptsize [CLS]}" at the beginning to comply with BERT. For action prediction, we assign the label of the next action to ``\texttt{\scriptsize [CLS]}". For query and variable prediction, we label the query and variable spans in the context using ``B-QUERY", ``I-QUERY", ``B-PARAMETER" and ``I-PARAMETER". It is worth noting that the context string differs between action, query and variable prediction, as query and variable predictions require additional {\em temporal context}, which is the content that the current prediction is associated with. For example, query prediction is associated with the API temporal context ``\texttt{\scriptsize [AGENT] find place}" and the second parameter of template ``\texttt{\scriptsize It is \{\} and \{\} away.}" is associated with the temporal context ``\texttt{\scriptsize [AGENT] It is \{\} and \{\} away [PARAM] v1 distance = 5 mi}". Capitalized words inside brackets are special tokens. For action prediction, temporal context is empty. Similar to BERT, we use ``\texttt{\scriptsize [SEP]}" as delimiter to concatenate the dialog context and temporal context, and then feed them into the BERT encoder. We optimize the model based on the cross-entropy loss over a label sized fully connected layer.

We use the example from Figure~\ref{dialogs} to show how we formulate the three categories of prediction into three sequence labeling problems. Assume we are at the fourth turn, and the agent is about to reply to the user, after receiving user utterance ``Is it the one across from Coffee Connection?". The next command the agent executed is: \textit{{\tt find\_place}(``Coffee Connection", v1\_latitude, v1\_longitude)}.

\subsubsection{Action prediction} The first problem is to predict the action that the agent will take. The dialog context string is:
\texttt{\scriptsize \underline{[CLS]} [USER] I want to go to Starbucks ... [AGENT] find\_place [VAR] v1 name = Starbucks [AGENT] There is a Starbucks ... [AGENT] Are you okay ... [USER] Is it ... from Coffee Connection ? [SEP]}
The temporal context is empty. We would assign the label ``B-find\_place" to ``[CLS]" token and label ``O" to all other tokens.

\subsubsection{Query prediction.} The second problem is to predict the query of \texttt{find\_place}. For this task, we append \texttt{\scriptsize [AGENT] find place} to the context (replacing underscore with space for API names): 
\texttt{\scriptsize [CLS] [USER] I want to go to Starbucks ... [AGENT] find\_place [VAR] v1 name = Starbucks [AGENT] There is a Starbucks ... [AGENT] Are you okay ... [USER] Is it ... from Coffee Connection ? [SEP] [AGENT] find place}.
Here we would expect label ``B-QUERY" to ``Coffee", and ``I-QUERY" to ``Connection". All other tokens would be labeled ``O".

\subsubsection{Parameter prediction.} Similarly, to predict the second parameter of \texttt{find\_place}, we would first append the query to the context, and also introduce the candidate variables \texttt{\scriptsize [VAR] source latitude | ...} and \texttt{\scriptsize [VAR] v1 latitude | ...} into the context:
\texttt{\scriptsize [CLS] [VAR] source latitude | name = source ... [USER] I want to go to Starbucks ... [AGENT] find\_place \underline{[VAR] v1 latitude | name = Starbucks,} \underline{neighborhood = Mar Vista ...} [AGENT] There is a Starbucks ... [AGENT] Are you okay ... [USER] Is it ... from Coffee Connection ? [SEP] [AGENT] find place [PARAM] Coffee Connection}.
Here, the underlined tokens are the correct parameter for the action. We would assign ``B-PARAMETER" to ``[VAR]" and ``I-PARAMETER" to the rest of the underlined tokens. 

Besides the source latitude and longitude, which are pre-set, all of the inserted variables come from invoked API calls. As the dialogs become longer, the number of candidates increases, and the machine learns to choose the correct variable. Instead of showing all variables to the machine, which may introduce noise, we only show variables whose types are valid. For instance: in the training set, the second parameter of \texttt{find\_place} is always latitude, so we only present latitude variables in the context. We obtain this parameter type information from the training data, for both API calls and templates. We do the same to predict the third parameter ``v1\_longitude".

The three prediction tasks share the same BERT encoder and fully connected layer at the top.

\begin{figure*}[t]
\begin{scriptsize}
\textbf{Context}
\begin{verbatim}
source_address = xxxx Admiralty Way, Marina del Rey
source_latlong = (33.9816425, -118.4409761)
USER: find a LA fitness near LAX
PREDICT: [ACTION] find_place [PARAM] LAX [PARAM] 33.9816425 [PARAM] -118.4409761
v1_address = 1 World Way, Los Angeles, CA 90045, United States
v1_name = Los Angeles International Airport
v1_latlong = (33.9415889, -118.40853)
v1_place_id = ChIJtU-yE9KwwoAR8a2LaVd7qHc
v1_street_number = 1
v1_street_name = World Way
v1_locality = Los Angeles
v1_distance = 4.7 mi
v1_duration = 17 mins
PREDICT: 
\end{verbatim}
\textbf{Generate}
\begin{verbatim}
[ACTION] find_place [PARAM] LA fitness [PARAM] (33.9415889, -118.40853)
\end{verbatim}
\end{scriptsize}
\caption{Representation of dialogs for training and deploying the GPT-based automatic agent.}
\label{GPT-prediction}
\end{figure*}

\subsection{GPT-based approach}

We fine-tune a GPT-2 model \cite{radford2019language} on our dialogs. GPT-2 can generate impressively realistic text, and can be fine-tuned to adapt it to a range of tasks, including dialog system \cite{zhang2019dialogpt}.\footnote{We use the default parameters and fine-tune from the pre-trained gpt2 model provided in the HuggingFace transformers library \cite{wolf2019huggingfaces}. This model has approximately 124 million parameters. Running on a single Nvidia Tesla P100 GPU this model averages 0.4 seconds per call, which corresponds to one agent API call or utterance. We fine-tune this model for 10 epochs on our training set, and select the model checkpoint with the highest click-level prediction accuracy on our validation set.}

We represent the dialogs in plain-text format, including all user utterances, agent utterances, API calls, and their returned variables. We fine-tune the model on the causal language modeling objective using this data. To produce the agent's action at a given turn, we provide as context the dialog so far, and generate the next line of text which corresponds to the agent's predicted action. An example is shown in Figure~\ref{GPT-prediction}.  We also make several extensions:

\begin{itemize}
    \item Generating variable names instead of values
    \item Replacing underscores in variable names with spaces, to better leverage GPT's pre-trained representations of word pieces.
    \item Using DialoGPT \cite{zhang2019dialogpt} as a starting point instead of GPT-2 base.
\end{itemize}

\subsection{Hand-built approach}

We constructed a hand-built agent out of rules and states derived from patterns in the collected training dialogs, in iterative fashion.  We emphasize that effective dialog collection (Section~\ref{collection}) made this agent possible. 
The agent keeps persistent state variables describing the current destination, query, and relative landmark.  

Its actions consist of (1) updating the query state, (2) issuing API calls and offering destination options based on the results, (3) retrieving a destination's attributes (such as open/closed, or rating), and (4) deciding to end the dialog and start driving.

Keyword-based rules determine which action to take.  Queries are extracted from user utterances using stopword lists, and they are updated with multi-turn context.  For example, after this dialog

\begin{itemize}
    \item User: ``I want to go to \textbf{Starbucks}.''
    \item Agent: ``I found a Starbucks on Lincoln Blvd, it is 3 min and 0.5 miles away.''
    \item User: ``I was thinking the one in \textbf{Culver City}.''
\end{itemize}

\noindent the query is updated to ``Starbucks Culver City''.  If a query is issued, the agent inserts result variables into simple templates such as ``I found {\em name} on {\em street} in {\em city}.  It is {\em minutes} and {\em miles} away.''  In case there is no update to the query, the agent responds with ``I'm sorry, I couldn't find anything.  Could you be more specific?''

\begin{figure*}[t]
\begin{center}
\begin{small}
\begin{tabular}{lrrrrr} 
\toprule
\multirow{3}{*}{\textbf{Method to predict agent behaviors}} & \multicolumn{4}{c}{\textbf{Click-level accuracy (\%)}} & \multirow{3}{*}{\textbf{\thead{Destination \\ accuracy}}} \\
\cmidrule(l){2-5}
& \textbf{overall} & \textbf{action} & \textbf{query} & \textbf{variable} & \\ 
& (n=1064) & (n=445) & (n=89) & (n=530) & \\ \hline
Human predictor & 68 & 31 & 92 & 70 & 100 \\ \hline 
Sequence-labeling (w/BERT) & 44 & 36 & 76 & 45 & 41 \\
\hspace{2mm} + parameter type constraint & 65 & 44 & 76 & 81 & 75 \\ 
\hspace{2mm} + include variables in action prediction context & 67 & 47 & 76 & 82 & 78 \\
\hspace{2mm} + restrict template parameters to only be variables & 68 & 47 & 76 & 85 & 79 \\ 
\hspace{2mm} + BERT large & 68 & 47 & 78 & 83 & 79 \\
\hspace{2mm} + agent embeddings & 70 & 50 & 73 & 85 & 81 \\
\hline 
LM prediction (w/GPT) & 64 & 46 & 74 & 78 & 83 \\ 
\hspace{2mm} + replace underscores with spaces & 63 & 39 & 75 & 80 & 79 \\
\hspace{4mm} + predict variable names & 68 & 45 & 63 & 88 & 98 \\
\hspace{2mm} + predict variable names & 65 & 41 & 73 & 84 & 90 \\
\bottomrule
\end{tabular}
\end{small}
\end{center}
\vspace{-0.1in}
\caption{Intrinsic results on held-out test set.  We record accuracy of predicting individual decisions of human agents.  These decisions include API selections, template selections, and parameter selections.  Destination accuracy reports variable accuracy of the final lat/long of the {\em start-driving} action (only).  The hand-built agent is not evaluated here, as it is not able (or required) to guess the ``next click'' performed by another (human) agent.}
\label{intrinsic-results}
\end{figure*}

\section{Evaluation}


{\bf Intrinsic.}  Figure~\ref{intrinsic-results} shows intrinsic results on our test set.  We include a human prediction baseline, in which a (separate) human tries to guess what individual decisions the (original) human agent took.  Human prediction is not 100\%, because multiple actions are reasonable.  


\begin{figure}[t]
    \centering
    \includegraphics[width=7.5cm]{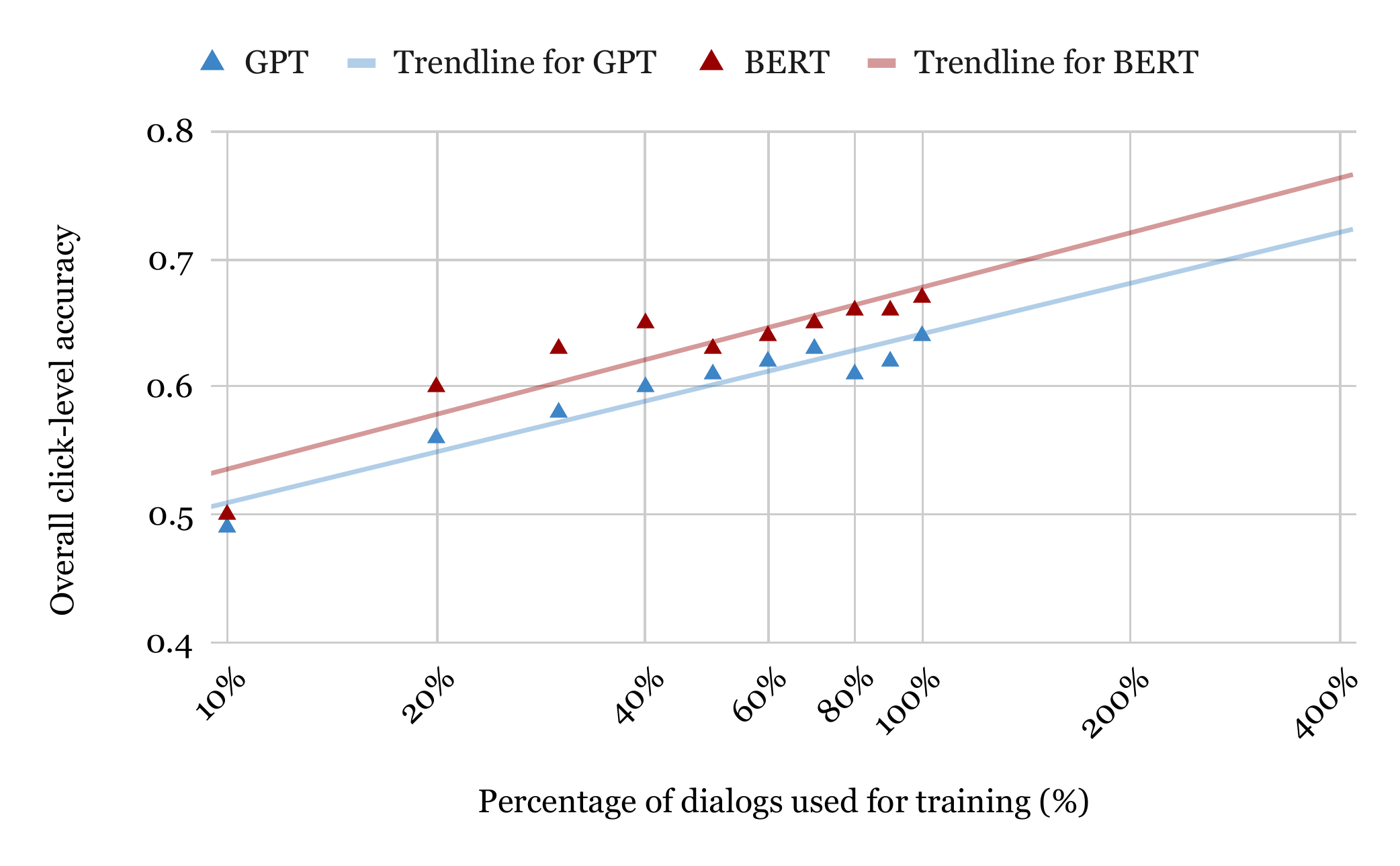}
    \caption{Learning curves for click-level accuracy.  The x-axis indicates amount of training data (log scale).  Extrapolation suggests more data would help accuracy. 
    }
\label{learning-curve}
\end{figure}

\begin{figure}
\begin{small}
\begin{center}
\begin{tabular}{lll} \hline
Source & Destination & First \\
address & description & utterance \\ \hline
{\em anon} & {\em anon} on {\em anon} Ave. & I'd like to go to a rock \\
& & climbing gym in {\em anon},\\
& & I forget the name. \\ \hline
{\em anon} & Donut store $<$10  & donuts please \\
& minutes away & \\ \hline
{\em anon} & Any karaoke studio, & find me a ktv \\
& preferably with a &  \\
& higher rating &  \\ \hline
{\em anon} & Bubble tea place, but & I'd like to \\
& not {\em anonymized} & get bubble tea! \\ \hline
\end{tabular}
\end{center}
\end{small}
\caption{Examples of the test cases used in extrinsic evaluation. For each test case, we provide the human user with the source address, the destination description, and the first utterance. 
This removes variation when we evaluate agents.}
\label{extrinsic-examples}
\end{figure}

\begin{figure}
\begin{footnotesize}
\begin{center}
\begin{tabular}{lrrr}
\toprule
\textbf{Method} & \textbf{Task}  & \textbf{Average} & \textbf{Average} \\
       & \textbf{compl. } & \textbf{\# of turns} & \textbf{\# of turns} \\
       & \textbf{rate} & \textbf{(success)} & \textbf{(fail)} \\ \hline
Human agent & 100.0\% & - & - \\ 
Hand-built agent & 90.1\% & 5.1 & 7.7 \\
BERT-based agent & 65.2\% & 5.2 & 7.6 \\ 
GPT-based agent & 57.4\% & 5.3 & 7.9 \\ \bottomrule
\end{tabular}
\end{center}
\end{footnotesize}
\caption{Extrinsic results. We record how often an automatic agent successfully guesses a destination across all human users assigned to it.}
\label{extrinsic-results}
\end{figure}

Our best trained system obtains 70\% click-level accuracy.  When we carry out subjective judgments on the system's decisions, we actually find that approximately 90\% are reasonable.  However, the 10\% mistakes are severe and generally result in the system going off track. 



Figure~\ref{learning-curve} extrapolates click-level accuracy with a learning curve. We subsequently collected an additional 104  dialogs to extend training. With a 20\% larger training set, we observe a 2\% absolute click-level accuracy gain for the BERT-based agent, showing that training corpus size is a bottleneck.

{\bf Extrinsic.}  We conduct an extrinsic evaluation to test how automatic agents perform in the real-world scenario. We let humans chat with an agent to specify a trip destination, and measure whether the agent successfully finds the correct latitude/longitude. To compare agents fairly, we devise 47 shared test cases. Each test case is a source address, a manually-written destination constraint, and a first user utterance. 
Figure~\ref{extrinsic-examples} shows examples of our test cases. A human user starts the conversation by sending the provided first utterance to the agent, and continues the dialog freely. 

Figure~\ref{extrinsic-results} shows extrinsic evaluation results. We recruit 12 volunteer evaluators: 3 for hand-built agent, 6 for BERT-based agent, and 3 for GPT-based agent. Each volunteer only evaluates one agent. Human agents operating in our GUI were able to correctly guess destinations 100\% of the time.  Our best hand-built dialog system obtains 90.1\%, while our best trained system obtains 65.2\%. The difference between hand-built and BERT-based agents is statistically significant (t-test, $p$=0.014), while the difference between BERT-based and GPT-based agents is not ($p$=0.22).

\section{Conclusion and Future Work}

We release an open-source platform (MEEP) for collecting task-oriented dialog corpora and training end-to-end agents.  We demonstrate the platform by creating a corpus of human/human trip destination dialogs, plus three automatic agents.  

We plan to extend the platform to host and learn from interactions between: (1) human users and automatic agents, (2) fixed, simulated users and automatic agents, and (3) automatic users and automatic agents that explore APIs and communication strategies together.


\bibliography{acl2020,dialog}

\begin{thebibliography}{30}
\expandafter\ifx\csname natexlab\endcsname\relax\def\natexlab#1{#1}\fi

\bibitem[{Asri et~al.(2017)Asri, Schulz, Sharma, Zumer, Harris, Fine, Mehrotra,
  and Suleman}]{dialog29}
Layla~El Asri, Hannes Schulz, Shikhar Sharma, Jeremie Zumer, Justin Harris,
  Emery Fine, Rahul Mehrotra, and Kaheer Suleman. 2017.
\newblock Frames: a corpus for adding memory to goal-oriented dialogue systems.
\newblock In \emph{Proc. SIGDIAL}.

\bibitem[{Bordes and Weston(2016)}]{dialog43}
Antoine Bordes and Jason Weston. 2016.
\newblock Learning end-to-end goal-oriented dialog.
\newblock In \emph{Proc. ICLR}.

\bibitem[{Brewer(2000)}]{eco}
M.~Brewer. 2000.
\newblock Research design and issues of validity.
\newblock In \emph{Handbook of Research Methods in Social and Personality
  Psychology}. Cambridge University Press.

\bibitem[{Budzianowski et~al.(2018)Budzianowski, Wen, Tseng, Casanueva, Ultes,
  Ramadan, and Gasic}]{dialog51}
Paweł Budzianowski, Tsung-Hsien Wen, Bo-Hsiang Tseng, Iñigo Casanueva, Stefan
  Ultes, Osman Ramadan, and Milica Gasic. 2018.
\newblock {MultiWOZ} - a large-scale multi-domain wizard-of-oz dataset for
  task-oriented dialogue modelling.
\newblock In \emph{Proc. EMNLP}.

\bibitem[{Devlin et~al.(2018)Devlin, Chang, Lee, and
  Toutanova}]{devlin2018bert}
Jacob Devlin, Ming-Wei Chang, Kenton Lee, and Kristina Toutanova. 2018.
\newblock {BERT}: Pre-training of deep bidirectional transformers for language
  understanding.
\newblock \emph{arXiv preprint arXiv:1810.04805}.

\bibitem[{Dhingra et~al.(2017)Dhingra, Li, Li, Gao, Chen, Ahmed, and
  Deng}]{Dhingra2017EndtoEndRL}
Bhuwan Dhingra, Lihong Li, Xiujun Li, Jianfeng Gao, Yun-Nung Chen, Faisal
  Ahmed, and Li~Deng. 2017.
\newblock End-to-end reinforcement learning of dialogue agents for information
  access.
\newblock In \emph{Proc. ACL}.

\bibitem[{Eric et~al.(2017)Eric, Krishnan, Charette, and
  Manning}]{Eric2017KeyValueRN}
Mihail Eric, Lakshmi Krishnan, Francois Charette, and Christopher~D. Manning.
  2017.
\newblock Key-value retrieval networks for task-oriented dialogue.
\newblock In \emph{Proc. SIGDIAL}.

\bibitem[{Espl{\`a}-Gomis et~al.(2019)Espl{\`a}-Gomis, Forcada,
  Ram{\'i}rez-S{\'a}nchez, and Hoang}]{EsplGomis2019ParaCrawlWP}
Miquel Espl{\`a}-Gomis, Mikel~L. Forcada, Gema Ram{\'i}rez-S{\'a}nchez, and
  Hieu Hoang. 2019.
\newblock Paracrawl: Web-scale parallel corpora for the languages of the {EU}.
\newblock In \emph{Proc. MT Summit}.

\bibitem[{Fraser and Gilbert(1991)}]{fraser1991simulating}
Norman~M Fraser and G~Nigel Gilbert. 1991.
\newblock Simulating speech systems.
\newblock \emph{Computer Speech \& Language}, 5(1):81--99.

\bibitem[{Hauptmann(1989)}]{hauptmann89}
A.~G. Hauptmann. 1989.
\newblock Speech and gestures for graphic image manipulation.
\newblock In \emph{Proc. CHI}.

\bibitem[{Henderson et~al.(2014)Henderson, Thomson, and Williams}]{dialog73}
Matthew Henderson, Blaise Thomson, and Jason~D. Williams. 2014.
\newblock The second dialog state tracking challenge.
\newblock In \emph{Proc. SIGDIAL}.

\bibitem[{Hutchins and Somers(1992)}]{Hutchins1992AnIT}
William~J. Hutchins and Harold~L. Somers. 1992.
\newblock \emph{An introduction to machine translation}.
\newblock Academic Press.

\bibitem[{Jurafsky and Martin(2018)}]{jurafsky18}
Daniel Jurafsky and James~H. Martin. 2018.
\newblock \emph{Speech and Language Processing}.
\newblock Chapter 24, Draft of September 23, 2018,
  https://web.stanford.edu/~jurafsky/slp3/24.pdf.

\bibitem[{Kelley(1984)}]{Kelley1984AnID}
J.~F. Kelley. 1984.
\newblock An iterative design methodology for user-friendly natural language
  office information applications.
\newblock \emph{ACM Trans. Inf. Syst.}, 2:26--41.

\bibitem[{Kerr(2020)}]{uberphone}
Dara Kerr. 2020.
\newblock Uber dials it back 20 years, bringing its ride service to feature
  phones.
\newblock \emph{CNET}.
\newblock February 13, 2020. https://cnet.co/3fPG7rC.

\bibitem[{Koehn et~al.(2007)Koehn, Hoang, Birch, Callison-Burch, Federico,
  Bertoldi, Cowan, Shen, Moran, Zens, Dyer, Bojar, Constantin, and
  Herbst}]{Koehn2007MosesOS}
Philipp Koehn, Hieu Hoang, Alexandra Birch, Chris Callison-Burch, Marcello
  Federico, Nicola Bertoldi, Brooke Cowan, Wade Shen, Christine Moran, Richard
  Zens, Chris Dyer, Ondrej Bojar, Alexandra Constantin, and Evan Herbst. 2007.
\newblock Moses: Open source toolkit for statistical machine translation.
\newblock In \emph{Proc. ACL}.

\bibitem[{Li et~al.(2017)Li, Chen, Li, Gao, and Çelikyilmaz}]{dialog26}
Xiujun Li, Yun-Nung Chen, Lihong Li, Jianfeng Gao, and Asli Çelikyilmaz. 2017.
\newblock End-to-end task-completion neural dialogue systems.
\newblock In \emph{Proc. IJCNLP}.

\bibitem[{Radford et~al.(2019)Radford, Wu, Child, Luan, Amodei, and
  Sutskever}]{radford2019language}
Alec Radford, Jeff Wu, Rewon Child, David Luan, Dario Amodei, and Ilya
  Sutskever. 2019.
\newblock Language models are unsupervised multitask learners.
\newblock In bit.ly/3dMAo3Z.

\bibitem[{Rastogi et~al.(2020)Rastogi, Zang, Sunkara, Gupta, and
  Khaitan}]{Rastogi2020TowardsSM}
Abhinav Rastogi, Xiaoxue Zang, Srinivas Sunkara, Raghav Gupta, and Pranav
  Khaitan. 2020.
\newblock Towards scalable multi-domain conversational agents: The
  schema-guided dialogue dataset.
\newblock In \emph{Proc. AAAI}.

\bibitem[{Rieser et~al.(2005)Rieser, Kruijff-Korbayov{\'a}, and
  Lemon}]{Rieser2005ACC}
Verena Rieser, Ivana Kruijff-Korbayov{\'a}, and Oliver Lemon. 2005.
\newblock A corpus collection and annotation framework for learning multimodal
  clarification strategies.
\newblock In \emph{Proc. SIGDIAL}.

\bibitem[{Shah et~al.(2018)Shah, Hakkani-Tür, Liu, and Tür}]{dialog12}
Pararth Shah, Dilek~Z. Hakkani-Tür, Bing Liu, and Gökhan Tür. 2018.
\newblock Bootstrapping a neural conversational agent with dialogue self-play,
  crowdsourcing and on-line reinforcement learning.
\newblock In \emph{Proc. NAACL-HLT}.

\bibitem[{Sutskever et~al.(2014)Sutskever, Vinyals, and
  Le}]{Sutskever2014SequenceTS}
Ilya Sutskever, Oriol Vinyals, and Quoc~V. Le. 2014.
\newblock Sequence to sequence learning with neural networks.
\newblock In \emph{Proc. NIPS}.

\bibitem[{Vaswani et~al.(2019)Vaswani, Shazeer, Parmar, Uszkoreit, Jones,
  Gomez, Kaiser, and Polosukhin}]{Vaswani2017AttentionIA}
Ashish Vaswani, Noam Shazeer, Niki Parmar, Jakob Uszkoreit, Llion Jones,
  Aidan~N. Gomez, Lukasz Kaiser, and Illia Polosukhin. 2019.
\newblock Attention is all you need.
\newblock In \emph{Proc. NIPS}.

\bibitem[{Wei et~al.(2018)Wei, Le, Dai, and Li}]{dialog8}
Wei Wei, Quoc~V. Le, Andrew~M. Dai, and Jia Li. 2018.
\newblock {AirDialogue}: An environment for goal-oriented dialogue research.
\newblock In \emph{Proc. EMNLP}.

\bibitem[{Wen et~al.(2017)Wen, Vandyke, Mrkšic, Gašic, Rojas-Barahona, Su,
  Ultes, and Young}]{dialog2}
Tsung-Hsien Wen, David Vandyke, Nikola Mrkšic, Milica Gašic, Lina~M.
  Rojas-Barahona, Pei-Hao Su, Stefan Ultes, and Steve Young. 2017.
\newblock A network-based end-to-end trainable task-oriented dialogue system.
\newblock In \emph{Proc. EACL}.

\bibitem[{Williams et~al.(2013)Williams, Raux, Ramachandran, and
  Black}]{dialog72}
Jason~D. Williams, Antoine Raux, Deepak Ramachandran, and Alan~W. Black. 2013.
\newblock The dialog state tracking challenge.
\newblock In \emph{Proc. SIGDIAL}.

\bibitem[{Williams and Zweig(2016)}]{Williams2016EndtoendLD}
Jason~D. Williams and Geoffrey Zweig. 2016.
\newblock End-to-end {LSTM}-based dialog control optimized with supervised and
  reinforcement learning.
\newblock \emph{arXiv preprint arXiv:1606.01269}.

\bibitem[{Wolf et~al.(2019)Wolf, Debut, Sanh, Chaumond, Delangue, Moi, Cistac,
  Rault, Louf, Funtowicz, and Brew}]{wolf2019huggingfaces}
Thomas Wolf, Lysandre Debut, Victor Sanh, Julien Chaumond, Clement Delangue,
  Anthony Moi, Pierric Cistac, Tim Rault, Rémi Louf, Morgan Funtowicz, and
  Jamie Brew. 2019.
\newblock \href {http://arxiv.org/abs/1910.03771} {Huggingface's transformers:
  State-of-the-art natural language processing}.

\bibitem[{Zhang et~al.(2019)Zhang, Sun, Galley, Chen, Brockett, Gao, Gao, Liu,
  and Dolan}]{zhang2019dialogpt}
Yizhe Zhang, Siqi Sun, Michel Galley, Yen-Chun Chen, Chris Brockett, Xiang Gao,
  Jianfeng Gao, Jingjing Liu, and Bill Dolan. 2019.
\newblock {DialoGPT}: Large-scale generative pre-training for conversational
  response generation.
\newblock \emph{arXiv preprint arXiv:1911.00536}.

\bibitem[{Zhao and Eskenazi(2016)}]{zhao16}
Tiancheng Zhao and Maxine Eskenazi. 2016.
\newblock Towards end-to-end learning for dialog state tracking and management
  using deep reinforcement learning.
\newblock In \emph{Proc. SIGDIAL}.

\end{thebibliography}
\bibliographystyle{acl_natbib}

\end{document}